\documentclass[10pt,twocolumn,letterpaper]{article}

\usepackage{cvpr}
\usepackage{times}
\usepackage{graphicx}
\usepackage{amsmath}
\usepackage{amssymb}
\usepackage{multirow}
\usepackage[export]{adjustbox} 
\usepackage{enumitem}

\usepackage[pagebackref=true,breaklinks=true,letterpaper=true,colorlinks,bookmarks=false]{hyperref}

\cvprfinalcopy 

\def\cvprPaperID{ 26   (CVS2017)} 

\ifcvprfinal\fi
\begin{document}

\title{Learning To Score Olympic Events}

\author{Paritosh Parmar and Brendan Tran Morris\\
University of Nevada, Las Vegas\\
{\tt\small parmap1@unlv.nevada.edu, brendan.morris@unlv.edu}
}

\maketitle
%
\begin{abstract}
Estimating action quality, the process of assigning a "score" to the execution of an action, is crucial in areas such as sports and health care.
Unlike action recognition, which has millions of examples to learn from, the action quality datasets that are currently available are small -- typically comprised of only a few hundred samples.  This work presents three frameworks for evaluating Olympic sports which utilize spatiotemporal features learned using 3D convolutional neural networks (C3D) and perform score regression with i) SVR, ii) LSTM, and iii) LSTM followed by SVR.  An efficient training mechanism for the limited data scenarios is presented for clip-based training with LSTM.  The proposed systems show significant improvement over existing quality assessment approaches on the task of predicting scores of Olympic events \{diving, vault, figure skating\}.
While the SVR-based frameworks yield better results, LSTM-based frameworks are more natural for describing an action and can be used for improvement feedback.
\end{abstract}
%
\section{Introduction}
Action quality assessment refers to how well a person performed an action. Automatic action quality assessment has applications in many fields like sports and health care.  For example, an injured player or someone with mobility impairments could perform exercise therapy on their own without the cost and inconvenience of a physical therapist while still getting feedback on performance and suggestions on how to improve.  
On the other hand, automated scoring systems could be used as a trusted impartial second opinion to avoid scoring scandals where the partiality of judges was questioned \cite{fs14scandal,olyscandals}; notably the 2002 pairs and 2014 women's Figure Skating (Winter Olympics) results.  

\begin{figure}[t]
\centering
  \includegraphics[width=180pt]{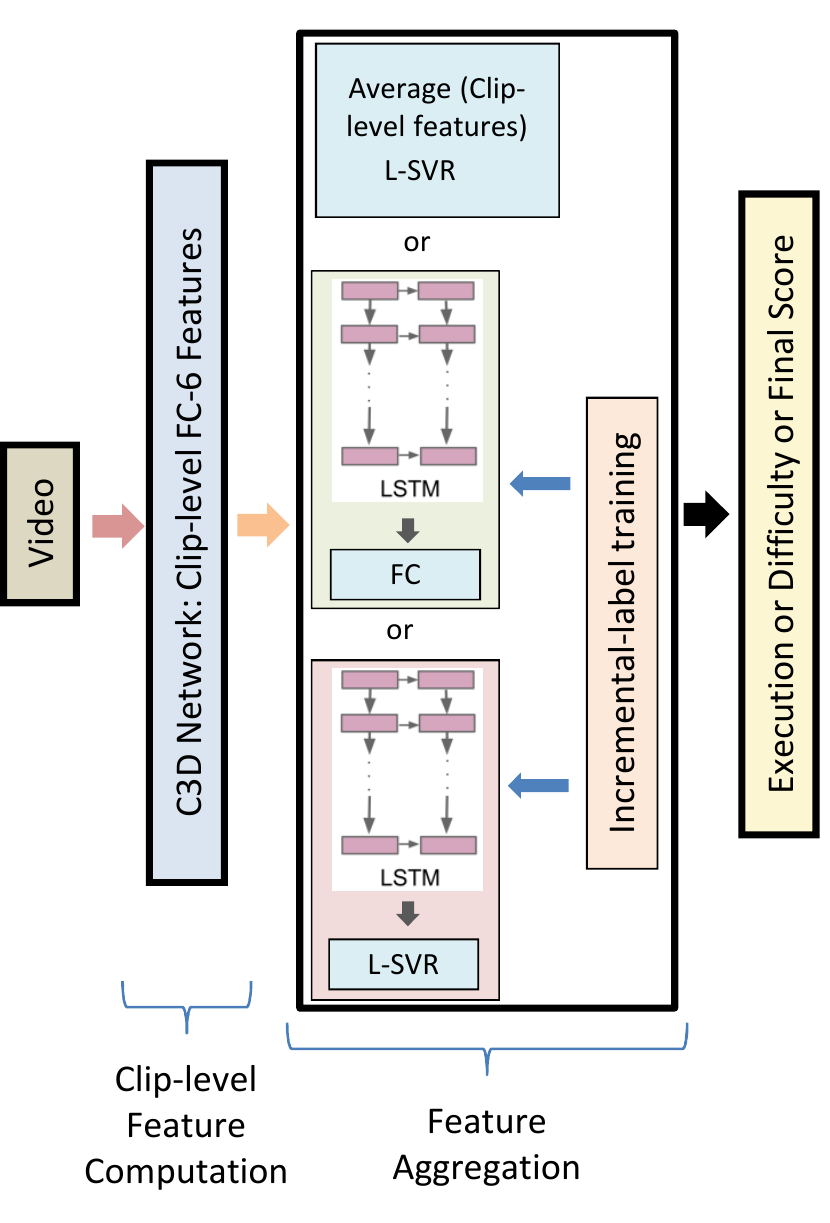}
\caption{Overview of proposed approaches.}
\label{fig:overview}
\end{figure}

Compared to action recognition, action quality assessment has received little attention. There are some key differences between action recognition and action quality measurement. Firstly, typically for an action recognition task, there is a significant amount of difference between two different classes of actions. In the case of action quality measurement, the difference between actions could be subtle similar to fine-grained classification. Secondly, while it has been shown that an action class can be recognized by ``seeing'' only a part of the action \cite{KarpathyCVPR14}, it is not meaningful to measure the quality of an action by seeing only a small part of the whole action. It is not meaningful because, there is a possibility that the performer will make an error at any given segment of the action. As an example, a diver may be perfect through the air but fail to enter the water vertically and make a large splash which is reflected as a poor overall dive score.  Therefore, if the dive was judged just by a short clip while the diver was in the air, the resulting perfect score would poorly correlate with the actual action quality.  However, if the whole action clip were taken into account, the diver would have had been penalized for erroneous entry into the water.

Convolutional neural networks, in particular the recently proposed 3D neural network (C3D) \cite{Tran_iicv2015} which learns spatiotemporal features, are increasingly being used for action recognition \cite{Tran_iicv2015, KarpathyCVPR14, varol16, Ng_2015_CVPR, Ye_2016_CVPR_Workshops, HengWang:2011:ARD:2191740.2192078}. Further, recurrent neural networks are used to capture the temporal evolution of a video \cite{Ng_2015_CVPR, Ye_2016_CVPR_Workshops}.  To train deep networks, large datasets are needed. For action recognition, many datasets such as UCF-101 \cite{ucf101}, HMDB51 \cite{Kuehne11} and Sports-1M \cite{KarpathyCVPR14} are available. Compared to action recognition, fewer action quality datasets \cite{pirsiavah_eccv2014, parmar_embc2016} are available and, in addition, action recognition datasets have millions of samples while the MIT Diving Quality dataset \cite{pirsiavah_eccv2014} contains just 159 samples.  Action recognition dataset can be increased in size by mining websites like YouTube using a script, whereas, increasing action quality dataset requires qualified human annotations to score the action, which makes it more labor intensive. 

\begin{figure*}
\centering
\includegraphics[width=0.8\linewidth]{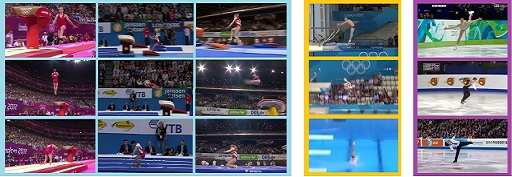}
   \caption{Images from gymnastic vault, diving and figure skating datasets. For gymnastic vault, we illustrate the viewpoint variations; first row shows the take-off, second row shows the flight while the third row shows the landing; images from different samples are shown in different columns.}
\label{fig:action_snaps}
\end{figure*}

When developing an action quality framework, it is critical to respect the constraint of small dataset size.  In this paper, we propose multiple frameworks that use visual information for action quality assessment and evaluate on short-time length action (diving and gym vault) and long-time length action (figure skating). Results show significant improvements over state-of-the-art \cite{pirsiavah_eccv2014, venkataraman_bmvc2015} for predicting the score of Olympic sports.  The major contribution of this work can be summarized as follows:
\begin{itemize}[noitemsep,nolistsep]
\item Introduction of new datasets for sports score assessment: The existing MIT diving dataset \cite{pirsiavah_eccv2014} is doubled from 159 samples to 370 examples. A new gymnastic vault dataset consisting of 176 samples (see Fig. \ref{fig:action_snaps}) has been collected. Datasets are available at \url{http://rtis.oit.unlv.edu/datasets.html}
\item We propose multiple approaches for action quality assessment which makes use of visual information directly unlike existing approaches which utilize noisy human pose information \cite{pirsiavah_eccv2014, venkataraman_bmvc2015} which is difficult to obtain in complicated athletic actions. 
\item An incremental LSTM training strategy is introduced for effective training when samples are limited with improved prediction quality and reduced training time by about 70\%.
\item We demonstrate how our LSTM-based approaches can be used to determine where action quality suffered which can be used to provide error feedback.
\end{itemize}
%
\section{Related Work}
Only a handful of works directly address the problem of action quality assessment \cite{pirsiavah_eccv2014, parmar_embc2016, venkataraman_bmvc2015, Zia2015, wnukS10}. Wnuk and Soatto introduced the FINA09 diving dataset, perform a pilot study,  and conclude that, temporal information, which is implicitly present in videos, is vital, and estimation of human pose using HoG features is reliable \cite{wnukS10}.  Human pose features have also been used in \cite{pirsiavah_eccv2014, venkataraman_bmvc2015} to assess action quality of Olympic sporting events. In \cite{pirsiavah_eccv2014}, a pose estimator is run on every relevant frame and concatenated to form a large action descriptor.  The descriptor is post-processed (DCT, DFT) into features which are used for estimating the parameters of a support vector regression (SVR) model to predict the event scores (quality). Venkataraman et. al \cite{venkataraman_bmvc2015}, using the estimated pose for each frame, calculate the approximate entropy features and concatenate them to get a high-dimensional feature vector. Their feature vector better encodes dynamical information than DCT. One of the drawbacks of using human poses as a feature is that the final results are affected by incorrectly estimated poses. 
%

Quality assessment has been performed in the health field as well.  Surgical skills were assessed using spatiotemporal interest points with frequency domain transformed HoG-HoF descriptors.  A final classifier is learned to identify a surgeon's skill level \cite{Zia2015}.  Similarly, quality assessment in physical therapy has been cast as a classification problem \cite{parmar_embc2016}.  In this work, 3D pose information from a Kinect is used to determine if a exercise repetition was "good" or "bad".  

%
%
Pose estimation has been shown to be challenging on diving and figure skating datasets \cite{pirsiavah_eccv2014} due to atypical body positions.  Pose only descriptors neglect important cues used in sport "execution" scoring such as splash size in diving.  Relative pose also does not reflect absolute positioning information which may be important for score (e.g. entry position of a dive).  Therefore, visual features such as C3D are expected to perform better.  Additionally, Olympic sports have clear rules for scoring which can be exploited to separate the type of dive from the execution quality of the dive.  These rules have not been addressed by current literature.  

\section{Approach}

Instead of using human pose information explicitly, the proposed systems leverage visual activity information to assess quality of actions (see Fig.\ref{fig:overview}).  Since quality of a sport performance is dependent not only on appearance but evolution over time, the first stage of the proposed sports assessment system extracts spatiotemporal features from video using the C3D network.  C3D has been shown to be effective at preserving temporal information in video and outperform 2D ConvNets  \cite{Tran_iicv2015}.  The features tend to capture appearance in the beginning of a clip and after the first few frames focuses on salient motion making them well suited for activity analysis.  After feature extraction, three different frameworks are proposed which differ in the way they aggregate clip-level features to obtain a video-level (or equivalently, sample-level) description. In one framework, SVR is built directly on clip-averaged C3D features; the second framework explicitly models the sequential nature of an action using an LSTM; and the final framework combines the LSTM with SVR. 
\subsection{C3D-SVR}
To classify a given instance of action, Tran et al \cite{Tran_iicv2015} divided the whole action video into clips of 16 frames and extracted C3D features for every clip.  A video-level action descriptor was obtained by averaging the clip-level features and used as input to a SVM to output predicted action class.  With this inspiration, the first variant on action quality assessment follows the same pipeline but replaces the SVM with a SVR (Fig. \ref{fig:overview}(top)).  The clip-level features are obtained from the FC6 layer of the C3D network.  The final feature vector is the normalized temporal clip average which is used as input to a SVR trained using the action score (quality value).  Note that through clip-level aggregation, the temporal evolution and timing of an action is lost.
\subsection{C3D-LSTM}
In the C3D-LSTM approach, clip-level C3D features are combined to model sequential (temporal) effects through an LSTM layer to generate a video-level description.  This technique was inspired by Ng et. al \cite{Ng_2015_CVPR} who used multilayer LSTM's for sequential feature aggregation of up to 120 frames  and Ye and Tian \cite{Ye_2016_CVPR_Workshops} who use LSTM to embed sequential information between consecutive C3D clips and dense trajectories \cite{HengWang:2011:ARD:2191740.2192078} . 

Specifically, the C3D FC-6 layer activations are used as input to an LSTM (in this work, there are two parallel LSTMs to encode the execution and difficulty scores respectively).  Each LSTM is followed by one fully connected regression layer to map the clip-evolved LSTM features to a score.  
Another advantage of C3D features, beyond spatiotemporal description, is that it provides a more compact video representation than frame-level CNN resulting in fewer "steps" to process a video.  
Take an action instance of 145 frames.  Using C3D features, and a temporal stride of 16 frames, we can represent the whole action instance using just $145/16 = 9$ time steps; whereas if we had used frame-level description, our action instance would have been a $145$ time steps long. Generally, longer sequences need multiple LSTM layers to model the temporal evolution, while shorter sequences can be modeled efficiently using a single layer LSTM. Therefore C3D features limits the number of LSTM parameters and hence the number of training samples required.
%
\begin{figure*}
\begin{center}
\includegraphics[width=0.95\linewidth]{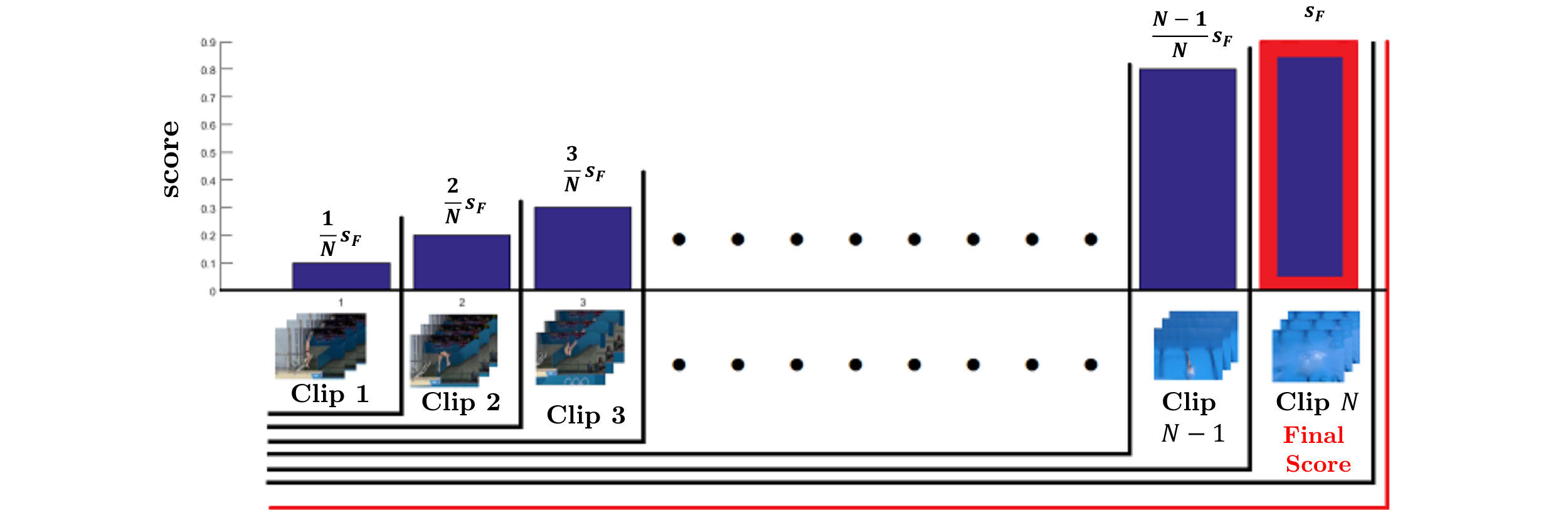}
\end{center}
   \caption{Difference between \textit{incremental-label training} (blue) and \textit{final-label training} (red).  In final-label training, the LSTM network sees all the clips in a video (accumulated observation from start to end of action) and the error with the final score is used for parameter updates through back-propagation.  With incremental-label training, back-propagation occurs after each clip through the use of an intermediate score which accounts for all frames observed to the current clip.  For simplicity, the intermediate score is a linear accumulation in time.  
}
\label{fig:lstm_training}
\end{figure*}
\subsubsection{LSTM Final-Label Training}
%
%
The problem of action quality assessment is at heart a many-to-one mapping problem, because, given a stack of frames (or equivalently clips), we want to predict a score (either execution score or difficulty score). 
We refer to training a LSTM on the single final label, i.e. score at the end of an event, as \textit{final-label training}.  In final-label training, all clips in a video are sequentially propagated through the LSTM network and the error is computed upon completion by comparison with the final video score $s_F$ (red in Fig. \ref{fig:lstm_training}). 

As the input for the LSTM, we use the FC-6 layer activations from C3D, and Euclidean distance as the loss function for training. A fixed learning rate policy of 0.0001 is used, input video is randomly cropped to $112\times112$ pixels, and input is randomly flipped as described in the C3D paper \cite{Tran_iicv2015}. 
%
%
%
%
There are two sets of unknowns that need to be determined during \textit{final-label training}. The first set of unknowns is the partial score attributed to different stages of an action. The second unknown is the total score after seeing all input clips. This will be a difficult task with limited size datasets ($<$400 examples).  
\subsubsection{LSTM Incremental-Label Training}
It is expected that as an action advances in time, the score should build up (if the quality is good enough) or be penalized (if the quality is sub par). For example, let's consider a two-somersault, one-and-half-twists dive. This dive can crudely be divided into various stages like diver's take-off, completing first somersault through air, completing second somersault through the air, completing all the twists, entering the water, etc. Each of these stages should have a small contribution in the final score. Following this intuition, score should be accumulated through an action as a non-decreasing function.  

To improve the LSTM training with limited data, we propose a new training protocol called \textit{incremental-label training}. In incremental-label training, unlike final-label training, we use intermediate labels when training LSTM's instead of using just the final label $s_F$ for the whole action instance. The intermediate label $s(c)$ for a given time step is supposed to indicate the score accumulated up until the end of clip $c$.  The concept of incremental scores is depicted in blue in Fig. \ref{fig:lstm_training}.  

The intermediate labels can be obtained in either supervised or unsupervised manner.  In the supervised case, an annotators must identify \textit{sub-actions} (smaller segments of an action) and the accumulated scores at each sub-action time.  Considering the previous diving example, sub-actions like somersault and twists have standardized scoring from the the FINA governing body, however, segmentation into the sub-action units could be difficult.  In practice it is not feasible to use supervised scores since it would require an expert judge to segment sub-actions, e.g. somersault and twist, and assess their worth using official FINA rules.  Ignoring effort, C3D clips are fixed frame length and these are not expected to align perfectly with sub-actions.  


In contrast, the unsupervised assignment can save time and effort and provide a scalable solution.  A video is divided into clips and the total score is evenly divided into each clip (each clip contributes the same amount of score).  The intermediate score for each clip is
\begin{equation}
s(c) = \frac{c}{N}\times s_F
\end{equation}
where $c$ is the clip number and $N$ is the number of clips in a video.  Incremental-label training is used to guide the LSTM during the training phase to the final score with intermediate outputs (i.e. back-propagation occurs after each clip).  Since the unsupervised assignment does not strictly respect the sub-action scores, in practice, we utilize a two step training process.  The LSTM is first trained for a few thousand iterations using incremental-label training.  The model is then fine-tuned by using final-label training at a lower learning rate.
The final-label fine-tuning works well in practice to relax the constant score growth restrictions from the incremental-labels. 

\subsection{C3D-LSTM-SVR}
In C3D-LSTM-SVR, the same C3D-LSTM network as discussed previously is employed.  The final fully connected regression layer after the LSTM is removed and we train a SVR, on LSTM layer activations, to predict action quality score (Fig. \ref{fig:overview} bottom).  This architecture provides explicit sequence and temporal modeling of an action through the LSTM while taking advantage of the shallow discriminative SVR for generalization in the face of limited data.
\subsection{Error detection}
\begin{figure*}[t]
\begin{center}
\includegraphics[width=0.8\linewidth]{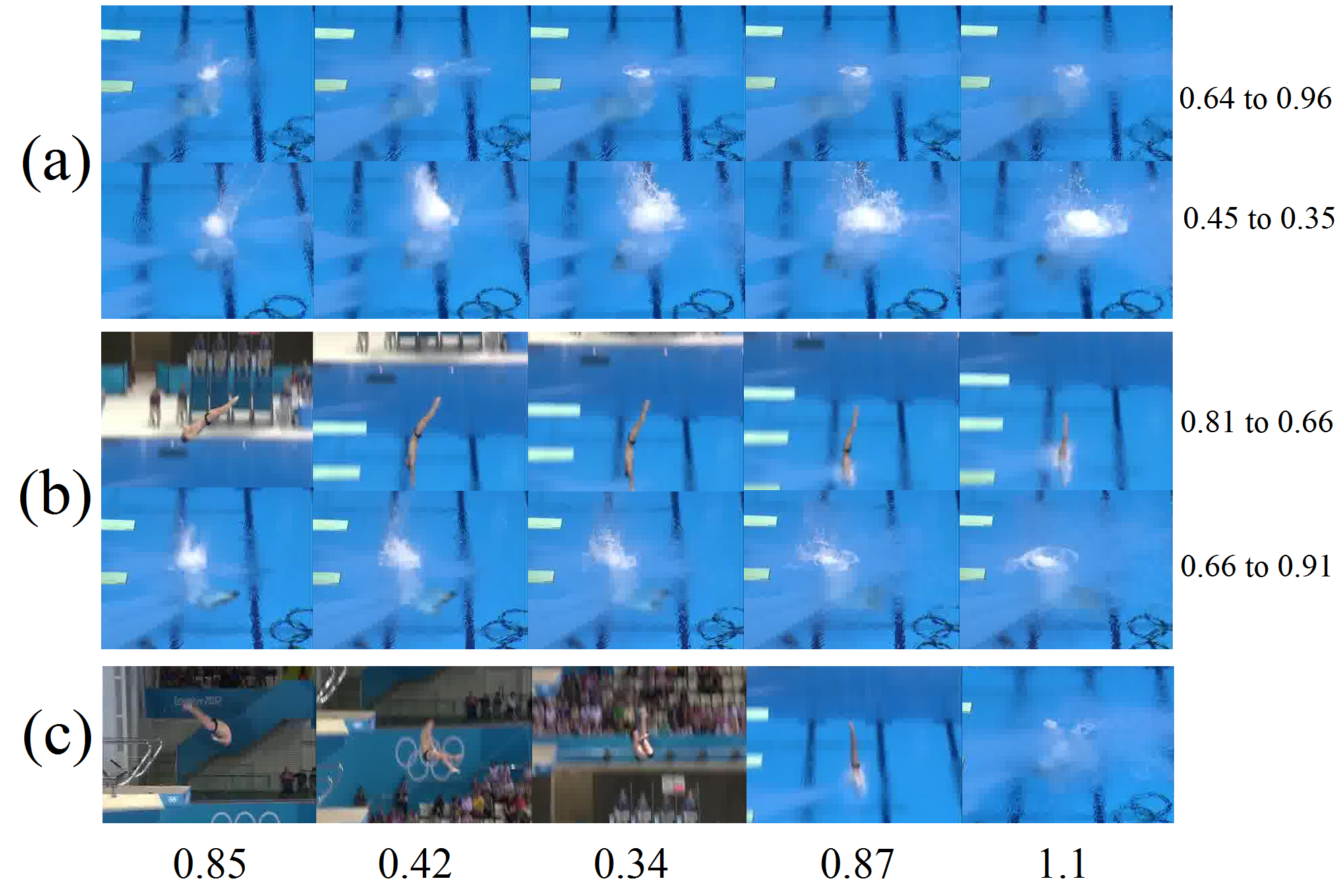}
\end{center}
   \caption{Error detection: (a) Top row: diver managing a 'ripping' -- one with almost no splash -- which resulted in normalized score increase from 0.64 to 0.96.  Bottom row: a dive with larger splash where score decreased. (b) Top row: dive with non-vertical entry with corresponding score decrease.  Bottom row: despite poor entry, only a small splash appears and is rewarded with increased score.  (c) Score evolution by clip: the diver has a strong take-off from the platform followed by legs split at ankles during the somersault in pike position which results in reduction from 0.85$\rightarrow$0.42$\rightarrow$0.34.  In the last clips there is a strong entry and little splash resulting in increasing score.}
\label{fig:feedback}
\end{figure*}

Intuitive and understandable spatiotemporal output is desirable in action quality assessment scenarios since this information could be presented to the human subject as a means to improve performance.  Pirsiavash et al. \cite{pirsiavah_eccv2014} generate a feedback proposal by first differentiating their scoring function with respect to joint locations, and then, by computing the maximum gradient, can find the joint and direction the subject must move to bring the most improvement in the score. In our case, we are not using human pose features, so we can not directly use such a feedback system. 
Instead, the temporal score evolution as it changes through the LSTM structure is utilized to identify both "good" and "poor" components of an action.  The assumption is that a perfectly executed action would have a non-decreasing accumulation of score while errors will result in a loss of score.  Error detection is illustrated in Fig. \ref{fig:feedback}. 
It should be noted though that the LSTM-feedback mechanism identifies the clip-level gain/loss but does not provide true explanation of why. 
%
%
\section{Experiments \& Results}

We evaluate the action quality assessment frameworks on three Olympic sports which are scored by judges, i) figure skating, ii) diving and iii) gymnastic vault. A full summary of all results can be found in Table. \ref{tab:overall}. 

\begin{table}[t]
\centering
\begin{tabular}{|l|c|c|c|c|}
\hline
\textbf{Dataset}                                                                 & \multicolumn{2}{c|}{\textbf{Diving}} & \textbf{Skating} & \textbf{\begin{tabular}[c]{@{}c@{}}Vault\end{tabular}} \\ \hline
\textbf{Samples}                                                                 & \textbf{100/59}   & \textbf{300/70}  & \textbf{100/70}  & \textbf{120/56}                                                    \\ \hline
\textbf{\begin{tabular}[c]{@{}c@{}}Pose+DCT {\cite{pirsiavah_eccv2014}}\end{tabular}} & 0.41              & 0.53             & 0.35             & 0.10                                                        \\ \hline
\textbf{\begin{tabular}[c]{@{}c@{}}ConvISA {\cite{le_cvpr2011}}\end{tabular}} & 0.19              & -             & 0.45             & -                                                        \\ \hline
\textbf{ApEnFT {\cite{venkataraman_bmvc2015}}}                                                                 & 0.45              & -                & -                & -                                                                  \\ \hline
\textbf{C-S}                                                                 & \textbf{0.74}     & \textbf{0.78}    & \textbf{0.53}    & \textbf{0.66}                                                      \\ \hline
\textbf{C-L (F)}                                                            & 0.05             & 0.01            & -                & -0.01                                                              \\ \hline
\textbf{C-L (I)}                                                            & 0.36              & 0.27            & -                & 0.05                                                               \\ \hline
\textbf{C-L-S (F)}                                                        & 0.56              & \textbf{0.66}    & -                & 0.33                                                               \\ \hline
\textbf{C-L-S (I)}                                                        & \textbf{0.57}     & \textbf{0.66}    & -                & \textbf{0.37}                                                      \\ \hline
\end{tabular}
\caption{Olympic score prediction comparison with literature. (C = C3D, S = SVR, L = LSTM, F = Final, I = Incremental). ConvISA results published in \cite{pirsiavah_eccv2014}.}
\label{tab:overall}
\end{table}

\textbf{Performance Metrics:} Action quality assessment is posed as a regression problem to predict the quality "score", as such, Spearman rank correlation $\rho$ is used to measure performance \cite{pirsiavah_eccv2014, venkataraman_bmvc2015}. Higher $\rho$ signifies better rank correlation between the true and predicted scores.  This metric allows for non-linear relationship, however, it does not explicitly emphasizes the true score value but relative ranking (i.e. lower scores for poor examples and higher scores for better quality examples). \\ 
\indent \textbf{Initial investigation:} An initial investigation was performed on a small diving dataset with 110 training and 82 validation samples.  Using 50-fold cross validation, it was found that FC6 of the full C3D architecture (pre-trained on Sports-1M) was the best layer for SVR regression (see Table \ref{tab:initial_investigation}) and the smaller C3D architecture (UCF-101) proposed by Tran et al. \cite{Tran_iicv2015} outperformed the full C3D.  Additionally, the C3D FC-6 features were sparse with 85\% zeros for small-C3D compared with 79\% with full-C3D.  Given improved sparsity and rank correlation, spatiotemporal features were extracted for video using the FC-6 activations from the small-C3D network in the following evaluations.  

\begin{table}
\centering
\begin{center}
\begin{tabular}{|l|c|l|l|l|}
\hline
\multicolumn{1}{|l|}{\textbf{Feature}} & \multicolumn{4}{l|}{\textbf{Correlation}} \\ \hline
conv5b (full architecture)                 & \multicolumn{4}{c|}{0.45}                  \\ \hline
pool5 (full architecture)                  & \multicolumn{4}{c|}{0.50}                  \\ \hline
fc6 (full architecture)                      & \multicolumn{4}{c|}{0.55}           \\ \hline
fc7 (full architecture)                      & \multicolumn{4}{c|}{0.46}                  \\ \hline
fc6 (small architecture)               & \multicolumn{4}{c|}{\textbf{0.63}} \\ \hline
\end{tabular}
\end{center}
\caption{Layer-wise correlation results.}
\label{tab:initial_investigation}
\end{table}

\subsection{Diving Dataset}
The original 10m platform diving dataset introduced in \cite{pirsiavah_eccv2014} (MIT-Dive) consisted of 159 samples which have been extend to a total of 370 samples by including dives from semi-final and final rounds of the 2012 Olympics (UNLV-Dive). A dive score is determined by the product of "execution", judged quality of a dive, multiplied by the dive "difficulty", fixed agreed-upon value based on dive type.  The execution score is in the range of [0, 30] in 0.5 increments whereas the difficulty has no explicit cap. The dive samples are recorded from a consistent side-view with little view variation.

The C3D features are obtained from training on UCF-101.  Evaluation utilizes two datasplits: i) 100 train / 59 test as in MIT Diving \cite{pirsiavah_eccv2014} for direct comparison to published results and ii) the extended 300/70 split to study dataset size effects.  

%
\subsubsection{C3D-SVR}

The C3D-SVR framework outperforms published results, Table \ref{tab:overall}.  Rank correlation is 80\% more than for Pose+DCT \cite{pirsiavah_eccv2014} and 65\%  better than the only other published results which used approximate entropy-based features (ApEnFT) \cite{venkataraman_bmvc2015} on the original MIT dive data.  In \cite{pirsiavah_eccv2014}, dive difficulty was unknown during evaluation, however, judges are provided this information for official scoring.  Table \ref{tab:divestride} highlights improved performance when C3D features are augmented with dive difficulty using the expanded UNLV-Dive dataset.  It was found that the optimal temporal stride was four and there was a $0.08$ correlation improvement.  Pose+DCT features were also augmented with difficulty but there was no improvement.  Further results will not utilize known difficulty to more closely approximate the MIT evaluation protocol.


\subsubsection{C3D-LSTM}
We implement our framework using Caffe \cite{jia2014caffe}.  All dive clips are padded with zero frames to length 151 to be the same length as the longest sample and split into 9 clips of 16 frames.  Two parallel LSTMs are implemented to predict the execution score separately from the difficulty score.  
%

\indent \textbf{Augmenting UCF-101:} We found that C3D-LSTM didn't work well with C3D network trained on the original UCF-101 dataset. UCF-101 contains diving as one of the 101 action classes. Upon visual inspection, the diving samples from UCF-101 are visually quite different from our diving samples. We believe that because of this difference, the C3D network is not able to give out very expressive description. To get better description from the C3D network, UCF-101 was augmented with our diving samples added to their diving class.  Additionally, during training, input videos were randomly flipped for improved performance.  
\begin{table}
\begin{center}
\begin{tabular}{|c|c|c|}
\hline
{\textbf{Stride}} & \textbf{- Difficulty} & \textbf{+ Difficulty} \\ \hline
1                                      & 0.77                              & 0.86                    \\ \hline
2                                      & 0.77                              & 0.86                               \\ \hline
4                                      & \textbf{0.78}                     & \textbf{0.86}                      \\ \hline
8                                      & 0.75                              & 0.81                               \\ \hline
16                                     & 0.74                              & 0.80                               \\ \hline
\end{tabular}
\end{center}
\caption{Result of varying ConvNet stride and inclusion of difficulty score on UNLV-Dive.}
\label{tab:divestride}
\end{table}


\textbf{Evaluation:} On the MIT-Dive data, training only required 1000 iterations for incremental-label training as compared with 10000 for final-label training.  Similarly, training was 5k and 18k for incremental- and final-label training on the UNLV-Dive data.  Note after incremental-label training, 2000 extra iterations of final-label training were performed to relax scoring constraints introduced through the linear incremental-score approximation.  Performance comparison between the two training methods is provided for both the execution-LSTM and the difficulty-LSTM on MIT-Dive on six different random train/test splits in Table \ref{tab:increvsfinal_159_overall}.  Incremental-training results much higher average $\rho=0.44$ versus final-label $\rho=0.14$ for execution score.  However, difficulty score was better with final-label training -- although, neither method worked well.  The overall score $\rho=0.36$ is less than Pose+DCT (Table \ref{tab:overall}). 

%
\begin{table*}[t]
\centering
\begin{tabular}{|c|c|c|c|c|c|c|c|c|c|c|c|c|}
\hline
               & \multicolumn{6}{c|}{\textbf{Incremental-label training (1K iterations)}}                           & \multicolumn{6}{c|}{\textbf{Final-label training (10K iterations)}}                                 \\ \hline
               & \multicolumn{3}{c|}{\textbf{C3D-LSTM}}           & \multicolumn{3}{c|}{\textbf{C3D-LSTM-SVR}}      & \multicolumn{3}{c|}{\textbf{C3D-LSTM}}           & \multicolumn{3}{c|}{\textbf{C3D-LSTM-SVR}}       \\ \hline
\textbf{Split} & \textbf{Exe}  & \textbf{Diff} & \textbf{Overall} & \textbf{Exe} & \textbf{Diff} & \textbf{Overall} & \textbf{Exe} & \textbf{Diff}  & \textbf{Overall} & \textbf{Exe}  & \textbf{Diff} & \textbf{Overall} \\ \hline
\textbf{1}     & 0.43          & -0.11         & 0.44             & 0.72         & 0.35          & 0.71             & 0.06         & 0.02           & 0.04             & 0.76          & 0.13          & 0.71             \\ \hline
\textbf{2}     & 0.38          & -0.18         & 0.37             & 0.56         & 0.02          & 0.48             & 0.18         & 0.04           & 0.29             & 0.64          & 0.26          & 0.44             \\ \hline
\textbf{3}     & 0.38          & -0.25         & 0.26             & 0.59         & 0.33          & 0.52             & 0.09         & 0.02           & -0.03            & 0.72          & -0.03         & 0.57             \\ \hline
\textbf{4}     & 0.52          & -0.07         & 0.52             & 0.63         & 0.19          & 0.60              & 0.15         & -0.10           & 0.00                & 0.71          & 0.07          & 0.64             \\ \hline
\textbf{5}     & 0.56          & -0.13         & 0.41             & 0.65         & 0.18          & 0.49             & 0.19         & 0.01           & 0.13             & 0.67          & 0.22          & 0.49             \\ \hline
\textbf{6}     & 0.39          & -0.38         & 0.18             & 0.74         & 0.26          & 0.62             & 0.18         & -0.04          & -0.16            & 0.76          & -0.08         & 0.54             \\ \hline
\textbf{AVG}   & \textbf{0.44} & -0.18         & \textbf{0.36}    & 0.65         & \textbf{0.22} & \textbf{0.57}    & 0.14         & \textbf{-0.01} & 0.05            & \textbf{0.71} & 0.09          & 0.56             \\ \hline
\textbf{STD}   & 0.07          & 0.11          & 0.12             & 0.07         & 0.12          & 0.09             & 0.05         & 0.05           & 0.15             & 0.05          & 0.13          & 0.10              \\ \hline
\end{tabular}
\caption{Rank Correlation on MIT-Dive Dataset performed over six random datasplits with results shown for Execution score, Difficulty score, and Overall Score.  Note incremental-training only required 1K iterations versus 10k for final-label.}
\label{tab:increvsfinal_159_overall}
\end{table*}

\begin{figure*}
\begin{center}
\includegraphics[width=0.7\linewidth]{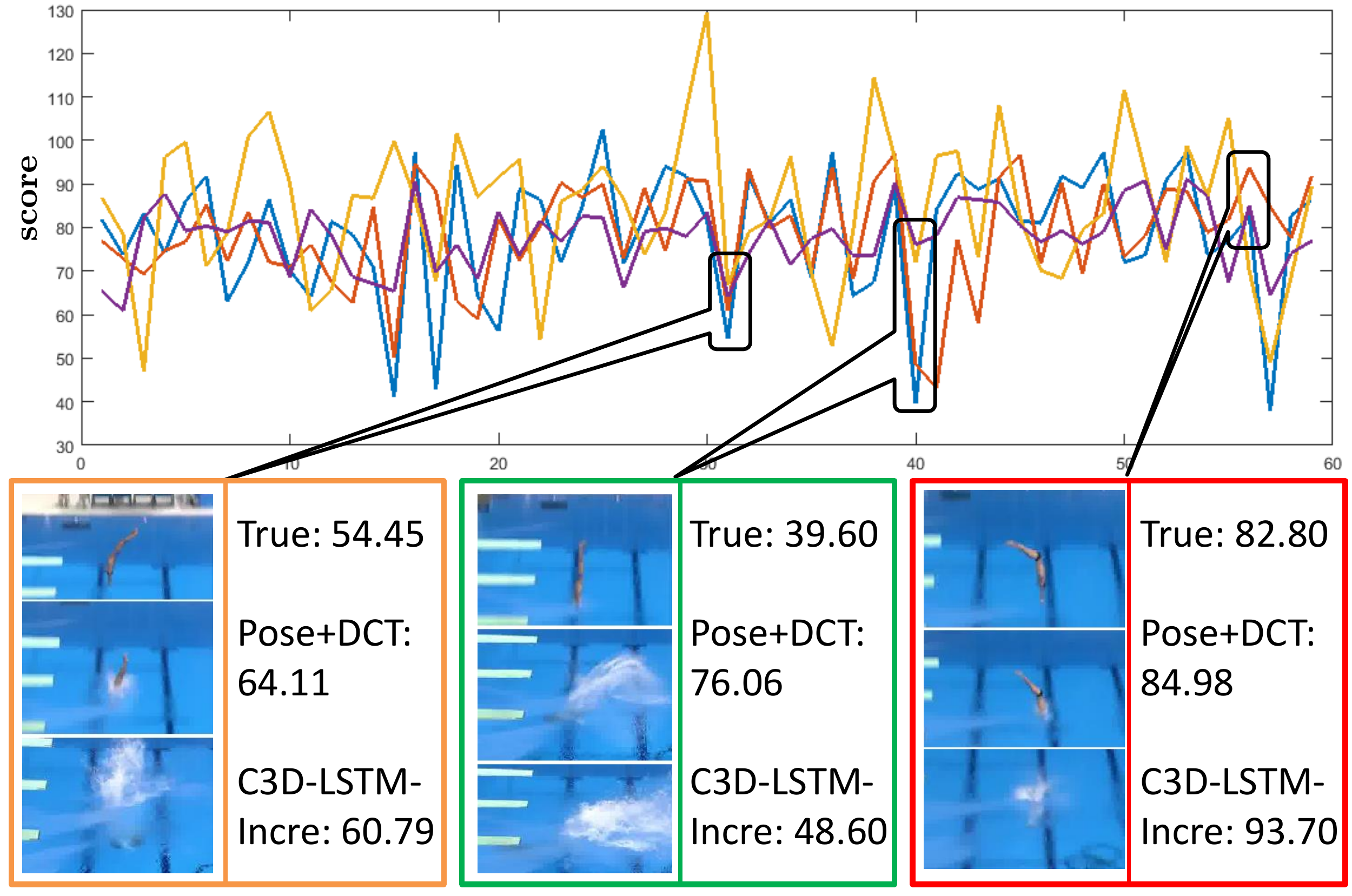}
\end{center}
\caption{Qualitative comparison among incremental-label training (red), final-label training (yellow) and Pose+DCT \cite{pirsiavah_eccv2014} (purple) and ground truth score (blue). Orange box: dive where all methods have similar output. Green box: C3D-LSTM-inc has much better prediction since it can recognize the splash whereas the pose only recognizes the clean entry body positioning and over-predicts.  Red box: Pose+DCT was better since it did not let the small splash outweigh the poor body position upon non-vertical entry into the water.}
\label{fig:qualitative_res}
\end{figure*}

\textbf{Analysis:} The LSTM implementation used 256 hidden nodes but did not find varying the number of nodes resulted in much improvement.  Adding a second LSTM layer did improve performance either, likely due to the added number of parameters to estimate on limited data.  Unlike with C3D-SVM above, using a denser sampling with stride of 8 instead of 16 reduced performance due to longer sequence dependency in a video (more clips).  Finally, the training data was augmented through temporal shifting.  With the 151 frames per dive, 7 end frames were not utilized when building 9 clips of length 16 frames.  Duplicate dives were created by shifting the start frame between frame 1 and 6 to augment UNLV-Dive from 300 to 1800 training samples.  Unexpectedly, the results were actually worse with more training data.

\subsubsection{C3D-LSTM-SVR}
Instead of averaging the clip-level C3D features, as we did in C3D-SVR, to get a video-level description, in C3D-LSTM-SVR, we use LSTM to model the temporal evolution. However, we take LSTM layer activations directly, without passing them through a fully-connected layer.  The SVR is used on top of the LSTM layer activations (as trained above). The results in Table \ref{tab:overall} clearly show improvements over the original C3D-LSTM.  Both incremental- and final-label training have similar results in this case at approximate gain of 40\% and 24\% on MIT- and UNLV-Dive respectively.  The SVR stage helps performance approach the C3D-SVM result with the added benefit of being able to due error detection (Fig. \ref{fig:feedback}) which could be used for feedback to the athlete for training or to explain judges scoring.

\subsubsection{Qualitative Comparison}
Score prediction results are shown qualitatively and compared against Pirsiavash et. al's approach in Fig. \ref{fig:qualitative_res}.  The plot gives the true score (blue), final-label (yellow), incremental-label (red), and Pose+DCT (purple) for the 59 test dives in MIT-Dive.  The orange box highlights dive 32 where all prediction methods obtained similar results which are close to the true score.  In the green box highlights dive 40 where Pose+DCT drastically over estimates the score since it only utilizes pose and does not recognize the large splash which the C3D-LSTM-incremental system is able track.  In the red box, Pose+DCT performs better since is explicitly accounts for the poor form while the C3D method seems to heavily weight the small splash.

\begin{table}
\begin{center}
\begin{tabular}{|l|c|}
\hline
\multicolumn{1}{|c|}{\textbf{Method}}                                                       & \textbf{Correlation}   \\ \hline
Pose+DCT \cite{pirsiavah_eccv2014}                                                             & 0.35          \\ \hline
Hierarchical ConvISA \cite{le_cvpr2011}                                                                     & 0.45          \\ \hline
\begin{tabular}[c]{@{}l@{}}C3D-SVR (- Deductions)\end{tabular} & \textbf{0.50} \\ \hline
\begin{tabular}[c]{@{}l@{}}C3D-SVR (+ Deductions)\end{tabular}     & \textbf{0.53} \\ \hline
\end{tabular}
\end{center}
\caption{Figure skating dataset comparison.  ConvISA results published in \cite{pirsiavah_eccv2014}.}
\label{tab:figskatres}
\end{table}

\subsection{Figure Skating}
The figure skating annotations (MIT-Skate) include the "presentation", "technical", and final scores. The scoring is points-based where a fixed technical base value is assigned to each element of a performance.  The presentation score evaluates the quality of an executed element based on an integer [-3, 3]  scale. The final score is a sum of technical and presentation scores. On an average, figure skating samples are 2.5 minutes (4500 frames) long and with continuous view variation during a performance.  Due to the long length of the event videos, only C3D-SVR was evaluated.  

\indent \textbf{C3D-SVR:} We applied the same approach as diving to figure skating dataset. We used 100 samples for training, and the remaining 71 samples for testing. We repeat the experiments 200 times with different random datasplits, as done by authors in \cite{pirsiavah_eccv2014}, and then average the results.  Results were poor with $\rho=0.22.$  Since UCF-101 does not have figure skating examples, the `Yo-Yo' action samples were replaced with skating samples.  C3D was trained on the augmented UCF-101 with much better results.  Table \ref{tab:figskatres} compares the performance with published results and shows the use of C3D features gives 0.05 improvement in correlation.  Like in dive, explicit information on deductions (e.g. modified jumps) improved performance. 


\subsection{Gymnastic Vault}
The UNLV-Vault dataset is a new activity quality dataset which consists of 176 samples.  In the new Olympic vaulting scoring system, the final score is the sum of a 10-point execution value plus a difficulty value.  Sequences are short with an average length of about 75 frames. Although sequence lengths are comparable with that diving dataset, view variation is quite large among the vault samples (at different events) due to broadcast configurations (see Fig. \ref{fig:action_snaps}) making it a more difficult dataset to score.

%
Since the UCF-101 dataset does not contain vault as one of its classes, again, `Yo-Yo' action samples were replaced with vault training samples for training the C3D network.  The vault evaluation only uses a single defined datasplit of 120 train and 56 test sample and are compiled in Table \ref{tab:overall}.  This dataset turned out to be challenging due to systematic view variation.   

%
%

Similar to dive results, the C3D-SVR variant had the best performance.  The Pose+DCT system did not work at all since pose consistency was poor.  
%
The training and testing protocol was the same for vault except that the number of frames was fixed to 100, resulting in LSTM sequence length of 6 (6 clips in a video).  Training required 1k and 10k iterations for incremental- and final-label training respectively.  C3D-LSTM performed very poorly with only $\rho=0.05$ for incremental-label.  However, the SVR addition was still able to improve performance by almost quadruple of Pose+DCT.  

%

Table \ref{tab:overall} makes the case for the use of C3D features for action quality assessment.  While he LSTM formulation has lower rank correlation, the ability to interpret the results is beneficial.  Future work will need to provide more insight spatially (what body parts are in error) as well as temporally.

\section{Conclusion}
We introduce a new Olympic vault dataset and present three frameworks for action quality assessment which improve upon published results: C3D-SVR, C3D-LSTM and C3D-LSTM-SVR. The frameworks mainly differ in the way they aggregate clip-level C3D features to get a video-level description. This video-level description is expressive about the quality of the action. We found that C3D-SVR gave best results, but was not able to detect errors made in the course of performing an action. We improve the performance of C3D-LSTM by using a SVR on top of it and although the performance of C3D-LSTM-SVR is lower than C3D-SVR, it has an advantage of being able to spot the erroneous segments of an action. 
{\small
\bibliographystyle{ieee}
\bibliography{egbib}
}

\end{document}